\documentclass[letterpaper]{article} 
\usepackage[draft]{arxiv_pkg/arxiv_aaai25}  
\usepackage{times}  
\usepackage{helvet}  
\usepackage{courier}  
\usepackage[hyphens]{url}  
\usepackage{graphicx} 
\usepackage{natbib}  
\usepackage{caption} 
\usepackage{algorithm}
\usepackage{algorithmic}

\usepackage{microtype}
\usepackage{graphicx}
\usepackage{subfigure}
\usepackage{booktabs} 

\usepackage{amsmath}
\usepackage{amssymb}
\usepackage{mathtools}
\usepackage{amsthm}
\usepackage{multirow}
\usepackage{makecell}
\usepackage{xspace}
\usepackage{enumitem}
\usepackage{tabularx}

\newcolumntype{Y}{>{\centering\arraybackslash}X}
\newcolumntype{C}[1]{>{\centering\arraybackslash}p{#1}}
\setlist[itemize]{noitemsep, topsep=1pt}

\theoremstyle{plain}

\theoremstyle{definition}

\theoremstyle{remark}

\newcommand{\userllm}{\textsc{User-LLM}\xspace}
\newcommand{\crossattention}{\textit{cross-attention}\xspace}
\newcommand{\softprompt}{\textit{soft-prompt}\xspace}
\newcommand{\perceiver}{\textit{Perceiver}\xspace}

\newcommand{\earlyfusion}{\textit{early fusion}\xspace}
\newcommand{\latefusion}{\textit{late fusion}\xspace}

\newcommand{\SubItem}[1]{
    {\setlength\itemindent{15pt} \item[-] #1}
}

\usepackage{newfloat}
\usepackage{listings}
\DeclareCaptionStyle{ruled}{labelfont=normalfont,labelsep=colon,strut=off} 
\floatstyle{ruled}
\newfloat{listing}{tb}{lst}{}
\floatname{listing}{Listing}
\title{\userllm: Efficient LLM Contextualization with User Embeddings}
\author{
  Lin Ning\textsuperscript{\rm 1}\equalcontrib,
  Luyang Liu\textsuperscript{\rm 1}\equalcontrib\corresponding,
  Jiaxing Wu\textsuperscript{\rm 1},
  Neo Wu\textsuperscript{\rm 1},
  Devora Berlowitz\textsuperscript{\rm 2},
  Sushant Prakash\textsuperscript{\rm 2},\\
  Bradley Green\textsuperscript{\rm 1},
  Shawn O'Banion\textsuperscript{\rm 1},
  Jun Xie\textsuperscript{\rm 1}
}
\affiliations {
    \textsuperscript{\rm 1}Google DeepMind, 
    \textsuperscript{\rm 2}Google \\[0.001cm]
}

\usepackage{bibentry}

\begin{document}

\maketitle

\begin{abstract}

Large language models (LLMs) have achieved remarkable success across various domains, but effectively incorporating complex and potentially noisy user timeline data into LLMs remains a challenge. Current approaches often involve translating user timelines into text descriptions before feeding them to LLMs, which can be inefficient and may not fully capture the nuances of user behavior. Inspired by how LLMs are effectively integrated with images through direct embeddings, we propose \userllm, a novel framework that leverages user embeddings to directly contextualize LLMs with user history interactions. These embeddings, generated by a user encoder pretrained using self-supervised learning on diverse user interactions, capture latent user behaviors and interests as well as their evolution over time. We integrate these user embeddings with LLMs through cross-attention, enabling LLMs to dynamically adapt their responses based on the context of a user's past actions and preferences.

Our approach achieves significant efficiency gains by representing user timelines directly as embeddings, leading to substantial inference speedups of up to 78.1X. Comprehensive experiments on MovieLens, Amazon Review, and Google Local Review datasets demonstrate that \userllm outperforms text-prompt-based contextualization on tasks requiring deep user understanding, with improvements of up to 16.33\%, particularly excelling on long sequences that capture subtle shifts in user behavior. Furthermore, the incorporation of Perceiver layers streamlines the integration between user encoders and LLMs, yielding additional computational savings.


\end{abstract}

\section{Introduction}

Large language models (LLMs) have revolutionized natural language processing (NLP) \cite{NEURIPS2020_1457c0d6, JMLR:v24:22-1144, OpenAI_GPT4_2023, touvron2023llama, anil2023palm, geminiteam2023gemini} and demonstrated remarkable success across various domains, including language generation, summarization \cite{liu2023learning, basyal2023text}, and question answering \cite{NEURIPS2022_8bb0d291, wei2023chainofthought}. This success has sparked significant interest in their potential to power the next generation of personalized AI agent - systems that understand and respond to individual needs and preferences. While LLMs have shown remarkable capabilities, their effectiveness often hinges on the quality and relevance of the data they are trained on. A key challenge lies in how to effectively incorporate the wealth of information contained within complex and potentially noisy user timeline data into these models. These timelines, spanning a wide range of interactions from textual input and search queries to media consumption, social media activity, location visits, etc., represent a rich source of user behavioral data. These data hold valuable insights crucial for deep user understanding and tailored user experience.

Existing approaches often translate user timelines into lengthy textual descriptions before feeding them into LLMs. This approach, while conceptually straightforward, leads to significantly computational overhead during inference due to the long context windows required to process these textual representation. Moreover, it may fail to capture the subtle nuances of user behavior and preferences essential for achieving truly personalized AI experiences.


\begin{figure}
    \centering
    \includegraphics[width=1.0\columnwidth]{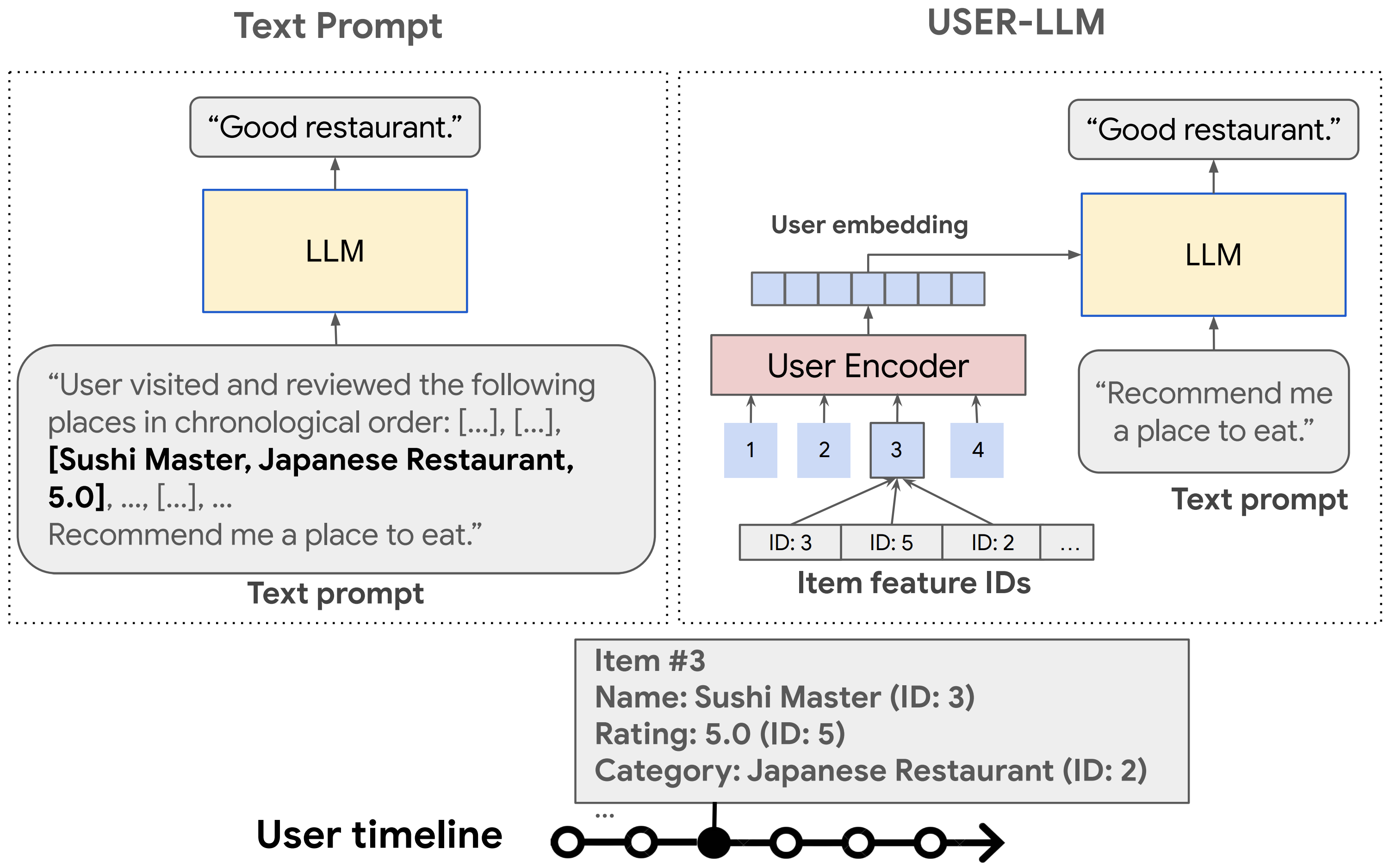}
    \caption{Illustration of two mechanisms to incorporate user interaction history data to LLMs. (1) Text Prompt: User history data are expressed in natural language and fed to LLMs via text prompt; (2) \userllm: User embeddings, distilled from user history data, are used to contextualize LLMs.}
    \label{fig:user-llm-motivation}
\end{figure}

Inspired by the effective integration of LLMs with other modalities, such as images \cite{alayrac2022flamingo,chen2022pali}, by directly incorporating embeddings from modality-specific encoders, we propose a paradigm shift for incorporating user history into LLMs. Instead of relying on computationally expensive textual representations, we treat user timelines as a distinct modality, directly embedding them for LLM contextualization.

This embedding-based approach offers several key advantages. First and most, it significantly reduces the context window length required for LLM processing, leading to substantial improvements in inference efficiency. Second, it allows LLMs to directly capture the latent patterns and temporal dynamics within user interactions, potentially fostering a deeper understanding of individual users and their evolving needs.


To realize this vision, we introduce \userllm, a novel framework (\ref{fig:user-llm-motivation}) that leverages user embeddings to directly contextualize LLMs. These embeddings, generated by a user encoder pretrained using self-supervised learning on diverse user interactions, encapsulate latent user behaviors, interests, and their evolution over time. We integrate these user embeddings with LLMs through cross-attention, enabling the models to dynamically adapt their responses based on the rich context of a user's past actions and preferences.

By representing user timelines directly as embeddings, we drastically reduce the computational burden on the LLM during inference, leading to inference speedups of up to 78.1X compared to text-prompt-based methods. Our comprehensive experiments across three datasets further demonstrate \userllm's ability to outperform text-prompt-based contextualization on tasks requiring deep user understanding, particularly with long context inputs.

Our \textbf{contributions} are four-fold:
\begin{itemize}
    \item We introduce \userllm, a framework that leverages user embeddings to directly contextualize LLMs, enabling them to dynamically adapt to a user's past actions and preferences. \userllm supports various user encoder architectures and encoder-LLM co-training strategies.
    \item We conduct extensive experiments on three public datasets, demonstrating that \userllm significantly outperforms text-prompt-based contextualization methods, achieving up to 78.1X inference speedup and up to 16.33\% improvement in performance on tasks requiring deep user understanding, particularly with long context inputs.
    \item We provide a comprehensive analysis of the impact of different user embedding generation architectures, encoder-LLM co-training strategies, and different ways to integrate embedding with LLMs on LLM personalization, offering valuable insights for future research.
    \item We delve into \crossattention mechanism for integrating user embeddings, exploring gated \crossattention to understand how user embeddings influence LLM behavior, and incorporate Perceiver layers for further efficiency gains.
\end{itemize}

\section{Related Work}

\subsection{Multimodal LLM}
Early work in multimodal LLMs primarily focuses on aligning images and text representations (e.g., CLIP~\cite{radford2021learning}, ImageBind~\cite{girdhar2023imagebind}). Subsequently, fusion techniques leveraging \crossattention (e.g., Flamingo~\cite{alayrac2022flamingo}, Coca~\cite{yu2205coca}) and \softprompt (e.g., PaLI~\cite{chen2022pali}, Palm-E~\cite{driess2023palm}) emerged. Recent works, such as NextGPT~\cite{wu2023next}, OneLLM~\cite{han2023onellm}, and Anymal~\cite{moon2023anymal}, explored unified frameworks for diverse input modalities. Additionally, end-to-end training of multimodal models has gained attention, as seen in Gato~\cite{reed2022generalist} and Gemini~\cite{team2023gemini}. Inspired by these advances, \userllm focuses on directly contextualizing LLM with embeddings for user understanding.

\subsection{Language Model based Personalization}

Recent research has extensively explored incorporating user interaction history as text prompts for LLM personalization and recommendation~\cite{petrov2023generative, kang2023llms,xu2023openp5,liu2023chatgpt, lyu2023llmrec, ji2023genrec, li2023exploring, wu2023survey}. However, directly feeding the entire user history can be computationally expensive  due to the resulting long context lengths. Our approach leverages user embeddings for efficient LLM contextualization, significantly reducing computational overhead. While a recent study \cite{Doddapaneni2024ArXiv} also explored LLM personalization with user embeddings, it focused on deriving embeddings from text-based user activity and integrating them via \softprompt. In contrast, \userllm utilizes efficient user activity tokens and sophisticated fusion techniques (e.g., \crossattention). We conduct a comprehensive evaluation across various datasets and tasks, demonstrating the effectiveness of \userllm. 

\subsection{Long Context in LLMs}
Long context is crucial for models to incorporate long-term user data. Existing approaches for LLMs include extending the context window (e.g., via positional encodings remapping \cite{chen2023extending, jin2024llm} or by exposing relevant additional contexts available to a subset of attention layers \cite{tworkowski2023focused}), distill information from the long context (e.g., by using a separate retrieval model \cite{xu2023retrieval}, a memory bank \cite{wang2023augmenting}, or training special tokens \cite{ge2023context, mu2023learning, chevalier2023adapting}), modified attention computation \cite{ding2023longnet, liu2023ring, chen2023longlora}, and linear attention frameworks (e.g., structured state space models \cite{DBLP:conf/iclr/GuJTRR23, gu2023mamba, 10.1145/3530811}). Our method tackles long context by using a single token to represent each user event, which is compatible with existing long context techniques.


\subsection{User Modeling}
Traditional approaches for user modeling and personalization have widely employed techniques like dual encoders, self-supervised learning~\cite{Yao2021self,xie2022contrastive,ChenLLMX22,XiaHHLYK23,Cai0XR23,yang2023debiased}, and pretrained models (e.g., BERT \cite{Jacob2018ArXiv}, Bert4Rec \cite{SunLWPLOJ19}, U-Bert \cite{Qiu_Wu_Gao_Fan_2021})) to generate effective user representations \cite{paul2016RecSys, covington2016deep, sountsov2016length, Volkovs2017NIPS, chidambaram2018learning, Gillick2018ArXiv, Yanga2018ACL, Ma2018KDD, Yi2019RecSys, gillick2019learning, Yang2020WWW, Jiang2020WWW, ning2021ArXiv}. While these approaches provide valuable insights into user behavior, they primarily operate outside the realm of LLMs. In contrast, our work focuses specifically on enhancing LLM capabilities for personalization by directly contextualizing them with user representations. This allows us to explore the potential of LLMs as the foundation for powerful, adaptive AI agents. Therefore, direct comparisons with non-LLM based user modeling techniques are not within the scope of this work.

\section{Methodology}
This section introduces the \userllm framework for contextualizing LLM with user embeddings (Fig.\ref{fig:user-llm}).

\subsection{Problem Statement}
Given a set of users $\mathcal{U}$ and a set of items $\mathcal{V}$, each user $u \in \mathcal{U}$ has a timeline sequence $S_u=(v_1^{(u)}, v_2^{(u)}, ..., v_{n_u}^{(u)})$ where each $v_i^{(u)} \in \mathcal{V}$ represents an item the user interacted with in chronological order. Each item $v$ is associated with a set of features $\mathcal{M}$. Each feature $m \in \mathcal{M}$ has a vocabulary $V_m$ that maps its values to integer IDs. For example, if items are movies, then the feature could be $\mathcal{M} = (name,\> genre,\> rating)$, where $V_{name}$, $V_{genre}$ map text values to IDs and $V_{rating}$ maps numerical ratings to IDs. Our goal is to develop a framework to contextualize a decode-only LLM with $S_u$, enabling it to generate personalized response $\mathit{A}_u$ for user $u$ given a query $\mathit{Q}$.

\subsection{\userllm}
\begin{figure*}[t]
\centering
    \includegraphics[width=0.85 \textwidth]{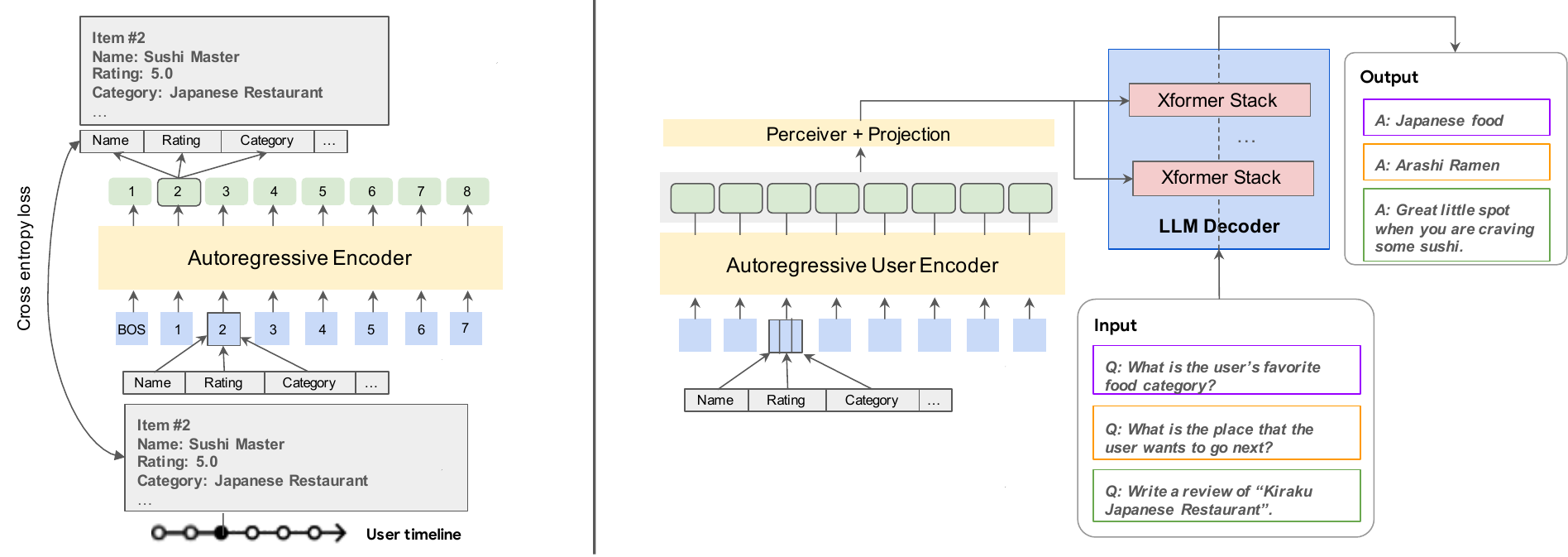}
    \caption{\textbf{Overview of \userllm}. (a) Autoregressive Transformer encoder pretraining. (b)LLM contextualization with user embeddings. Features from a user's interaction history are encoded by the Autoregressive User Encoder and then integrated into the language model via \crossattention.}
    \label{fig:user-llm}
\end{figure*}

As shown in Fig.\ref{fig:user-llm} (b), \userllm contains an ID-based user encoder and a decoder-only LLM. The {\bf inference} includes two-stages: (1) the user encoder generates user embeddings; (2) the embeddings are integrated into the LLM through \crossattention, providing the LLM additional context and guidance to generate personalized responses.

\subsubsection{User Embedding Generation}
\label{sec:embedding-gen}
\userllm leverages a Transformer-based encoder to generate user embeddings from ID-based interaction sequences. 

Let us consider a scenario where each item has $M$ features ($|\mathcal{M}| = M$). Representing each item with feature IDs, an ID-based user timeline sequence contains $\mathit{L}$ items can be denoted as
$$s_u = [(x_{1,m_1}, ..., x_{1,m_M}), ..., (x_{L,m_1}, x_{L,m_M})]$$
where $x_{i,m_j} = V_{m_j}[m_j^{v_i}]$ denotes the ID of feature $m_j$ for item $v_i$.

Each element ID has its unique embedding representation. To obtain a unified representation for an item, we combine embeddings from different features into a single embedding. Let $E_{m_j} \in \mathbb{R}^{|V_{m_j}| \times d_j}$ be the embedding matrix associated with $V_{m_j}$, where $d_j$ is the embedding dimension for $m_j$. For item $v_i$, the fused embedding is $f_i = [E_{m_1}(x_{i, m_1}); E_{m_2}(x_{i, m_2}); ..., E_{m_M}(x_{i, m_M})]$, where $[\cdot; \cdot]$ denotes concatenation. To reduce the dimension of the fused embedding, we project $f_i$ to $f_i'$ with a dimension of $d$.

\paragraph{\textbf{Autoregressive User Encoder}} As illustrated in Fig.~\ref{fig:user-llm} (a), an Autoregressive Transformer, denoted as $\mathcal{T}$, takes the fused embeddings $F' = [f_1', f_2', ..., f_L']$ into the Transformer decoder, and outputs embeddings $E_{s_u} = \mathcal{T}(F')$, where $E_{s_u} \in \mathbb{R}^{L \times d}$. These embeddings are used for LLM contextualization in the next stage.

Our framework is adaptable, supporting various user encoders. We've experimented with alternatives like Dual Encoders and achieved competitive results.

\subsubsection{LLM Contextualization with User Embedding}
\userllm integrates user embeddings with LLMs using \crossattention (see Fig.~\ref{fig:user-llm} (b)). The output embeddings from the user encoder are down sampled by a projection layer, and then are cross-attended with intermediate text representations within the LLM, similar to how Flamingo~\cite{alayrac2022flamingo} works. 

Let $E_{su}$ denote the encoder output, and $CrossAttn(Q, K, V)$ is the \crossattention mechanism. Note that for \crossattention, $Q$ is LLM input, whereas $K$ and $V$ are compatible with the dimension of cross-input $E_{su}$. Let $O_i$ and $I_{i+1}$ denote the output of the $i$-th and the input of $(i+1)$-th self-attention layers, respectively. Our framework enables \crossattention as follows:
\begin{align}
O_i &= \text{LayerNorm}(O_i) \label{eq:1} \\
O_i &= \text{CrossAttn}(O_i, E_{s_u}, E_{s_u}) + O_i \label{eq:2}\\
I_{i+1} &= \text{FeedForward} (O_i)
\end{align}
In order to further investigate the \crossattention mechanisms between encoders and LLMs, we also implemented gated \crossattention in \userllm by inserting gated \crossattention-only layers between LLM's self-attention stack. Flamingo-style gating mechanism \cite{alayrac2022flamingo} was employed to control the contribution of \crossattention signal to LLM. That is, a scalar between 0 and 1 is introduced to equation \eqref{eq:2}:
\begin{align}
    O_i &= \boldsymbol{\tanh(\alpha_i)} \times \text{CrossAttn}(O_i, E_{s_u}, E_{s_u}) + O_i 
\end{align}

where $\alpha_i$ is a trainable, zero-initialized scalar gating the $i$-th \crossattention layer. Using this approach, we enabled \crossattention between encoder output and  transformer layers in the LLM with layer-specific \crossattention weights regulated by $\alpha$. 

Note that our framework is adaptable to different integration mechanisms including prepending embeddings as \softprompt~\cite{lester-etal-2021-power}.

\subsection{Training Framework}
To effectively extract information from user interaction data and teach LLM to understand the information encoded in the user embeddings, we propose a two-stage training framework: user-encoder pretraining and encoder-LLM finetuning. 

\subsubsection{User-encoder Pretraining}
 In the first stage, we pretrain the Autoregressive Transformer on user interaction sequence so that it learns how to extract information from the input data. As described in Section~\ref{sec:embedding-gen}, the Autoregressive Transformer takes in a sequence of ID-based user interactions and outputs embeddings $E_{s_u}$. For pretraining, the decoder output is projected back to the original feature dimensions using feature-specific projection matrices $W_{m_1} \in \mathbb{R}^{d \times d_1}, W_{m_2} \in \mathbb{R}^{d \times d_2}, ..., W_{m_i} \in \mathbb{R}^{d \times d_i}$: $h_{m_j, i} = W_{m_j} e_i$, where $e_i \in E$ is the $i$-th embedding output from the Transformer.

\paragraph{\textbf{Loss Calculation}} The cross-entropy loss for each feature is calculated using the logits of the feature-specific projection:
$$\mathcal{L}_{m_j} = -\frac{1}{L}\sum_{i=1}^{L} \log (\text{softmax}(h_{m_j, i}))(f_{m_j, i+1}')$$
where $f_{m_j, i+1}'$ is the projected fused embedding for feature $m_j$ at step $i+1$. The overall loss is the average of cross-entropy losses across features: $\mathcal{L} = \frac{1}{i} \sum_{j=1}^{i} \mathcal{L}_{m_j}$


\subsubsection{Encoder-LLM Finetuning}
\label{sec:training_strategies}
In the second stage, we connected pretrained Autoregressive Transformer and a decode-only LLM with \crossattention and finetune the entire \userllm model to align user embeddings and LLM's text representations.
The model input has two parts: an ID-based user interaction sequence as the input to the Autoregressive Transformer ($X_{\mathcal{T}} = s_u$) and a query as the input to the LLM ($X_{LLM} = \mathit{Q}$).

\userllm offers flexible training strategies for Encoder-LLM finetuning to tailor its performance for diverse use cases. To thoroughly explore \userllm's capabilities, we investigated four strategies targeting different model components:
\begin{itemize}
\itemsep0.3em
    \item[$\circ$] \textbf{Full}: Finetuning the entire model (LLM, user encoder, and projection layers) enables maximum parameter adaptation to user interactions, revealing the upper bounds of \userllm's personalization potential.
    \item[$\circ$] \textbf{Enc}: Finetuning the user encoder and projection layers (LLM frozen) offers an efficient LLM contextualization approach while preventing the LLM from overfitting to specific user data. This maintains the LLM's general language capabilities.
    \item[$\circ$] \textbf{LoRA}: Finetuning the LLM with LoRA~\cite{hu2022lora}, along with the user encoder and projection layers, offers parameter efficiency preventing catastrophic forgetting and maintaining the LLM's core knowledge.
    \item[$\circ$] \textbf{Proj}: Finetuning only the projection layers (LLM and user encoder frozen) assesses the minimal parameter tuning required for effective LLM contextualization.
\end{itemize} 

\subsection{Efficiency}
Compared to text-prompt-based methods, \userllm offers substantial efficiency gains. First, it leverages pretrained weights of both the user encoder and LLM, improving model convergence speed with less trainable parameters. Additionally, \userllm condenses user activities into dense representations, requiring only a single token per event. This frees up the LLM's context window, leading to improved inference efficiency. Furthermore, \userllm leverages \crossattention, maintaining user embeddings at a smaller dimension than the LLM's model dimension, which significantly reduces computation during both training and inference.

To further optimize inference efficiency, \userllm integrates \perceiver~\cite{jaegle2021perceiver} units into its projection layers. \perceiver is a Transformer-based architecture that utilizes a trainable latent query to extract relevant information from input embeddings through \crossattention. In \userllm, \perceiver effectively compresses the user embeddings into a compact format using a latent query with shorter sequence length. This compression reduces the number of tokens needed to represent the user history, further freeing up the LLM's context window. This, along with \perceiver's ability to distill insights from noisy contexts, makes \userllm ideal for practical applications with extensive user histories.

\section{Experiments}
\label{sec:experiments}
\subsection{Datasets and Tasks}
We evaluated our approach on three widely recognized datasets: MovieLens20M~\cite{Harper2015MovieLens}, Amazon Review~\cite{amazon_data_2016}, and Google Review~\cite{Li2022ACL,Yan2023SIGIR}. Notably, Amazon Review dataset is well known for its high sparsity and variability and contains a significantly smaller number of training and evaluation examples.

We generated training and test examples by applying a sliding window over each user's feature sequences (sorted by timestamps). This creates inputs containing N items, with the subsequent item as the label. To ensure model evaluation consistency, the final example (representing the most recent interaction) is used as the test example for each user and is never included in the training set. Table~\ref{tab:dataset_info} summarizes datasets statistics after processing. 

To address the diverse evaluation needs, we accessed model performance on three distinct types of {\em open-text-generation} tasks. These tasks serve as a proof-of-concept, demonstrating \userllm's ability to understand user preferences from historical interactions and provide peronslized response accordingly.
\begin{itemize}
\itemsep0.3em 
    \item[$\circ$] \textbf{Next item prediction } Given a historical sequence of items, the model predicts the subsequent item. For example, predicting the next movie a user watches based on a sequence of previously viewed movie IDs.
    \item[$\circ$] \textbf{Favorite genre or category prediction } The model predicts a user's favorite genre or category based on a sequence of items (e.g., predicting favorite movie genre given a sequence of movie IDs). The favorite is the one with the highest frequency in the user's history. 
    \item[$\circ$] \textbf{Review generation } This task extends beyond simple predictions. Here, the model must generate actual reviews, utilizing multiple input features (e.g., movie ID, genre, and rating). These features are encoded to generate embeddings that guide the generation of a user's potential review for a given item.
\end{itemize}

\begin{table}[t]
    \centering
    \small
    \begin{tabular}{lcccc}
        \toprule
        \textbf{Datasets} & \textbf{\#Users} & \textbf{\#Items} & \textbf{\#Train} & \textbf{Seq Len} \\
        \midrule
        \scriptsize{MovieLens20M} & 82,977 & 27,280 & 13,821,405 & 50 \\
        \midrule
        \scriptsize{Google Review} & 80,080 & 270,721 & 3,387,375 & 50 \\
        \midrule
        \scriptsize{Amazon Review} & 5,756 & 177,978 & 357,258 & 50 \\
        \bottomrule
    \end{tabular}
    \caption{Dataset statistics. \#Test is equal to \#Users.}
    \label{tab:dataset_info}
\end{table}

\subsection{Baselines and Experiment Setup}
Our experiments used an Autoregressive Transformer for embedding generation.
For co-training, we explored four different training strategies described in Section~\ref{sec:training_strategies} and evaluated the impact of using a pretrained encoder versus a randomly initialized one.


We utilized PaLM-2 XXS~\cite{anil2023palm} as the pretrained LLM. The user encoder is a $6$-layer $128$-dimensional transformer with $8$ heads. We constructed the \textit{Perceivers} module with $6$ cross-attention layers and a query dimension of $16$. We used $rank=8$ for all LoRA tuning experiments.
To assess the effectiveness of \userllm, we compare it against the following two baselines:
\begin{itemize}
    \item[$\circ$] {\bf Text Prompt (TP)} contextualize LLM by finetuning it with raw text input derived from a user's timeline.
    \item[$\circ$] {\bf Text Summarization (TS)} contextualize LLM by finetuning it with text Summarization derived from a user's timeline. In particular, we first use a Gemini-M model to generate summarizations for raw text inputs of user history, and then finetune the PaLM-2 XXS towards different tasks. This is a time consuming process, and due to resource limitation, we only conduct comparison on Amazon Review datasets.
\end{itemize}

\subsection{Performance and Efficiency Comparison}
\label{subsec:performance-efficiency-comp}

\begin{table*}[t]
    \centering
    \small
    \begin{tabular}{ccccccc}
        \toprule
        \textbf{Dataset} & \textbf{Task} & \textbf{Metrics} & \textbf{Text-Prompt} & \textbf{User-LLM} & \textbf{Relative Improvement} & \textbf{FLOPs Reduction} \\
        \midrule
        \multirow{2}{*}{\makecell{MovieLens\\ 20M}}
        & NextItem & Recall@5 & 0.140 & 0.154 & 10.00\% & 21.9X \\ 
        \cmidrule{2-7}
        & FavGenre & Accuracy &  0.787 & 0.787 & 0.00\% & 21.9X \\ 
        \midrule
        \multirow{3}{*}{\makecell{Google\\ Review}} 
        & NextItem & Recall@5 & 0.057 & 0.052 & -8.77\% & 12X \\
        \cmidrule{2-7}
        & FavCategory & Accuracy & 0.741 & 0.862 & 16.33\% & 12X \\
        \cmidrule{2-7}
        & ReviewGen & Rouge & 10.20 & 11.72 & 14.90\% & 16X \\
        \midrule
        \multirow{3}{*}{\makecell{Amazon\\ Review}} 
        & NextItem & Recall@5 & 0.042 & 0.047 & 11.90\% & 16X \\
        \cmidrule{2-7}
        & FavCategory & Accuracy & 0.885 & 0.890 & 0.57\% & 16X \\
         \cmidrule{2-7}
         & ReviewGen & Rouge & 22.82 & 26.38 & 15.59\% & 16X \\
        \bottomrule
    \end{tabular}
    \caption{User-LLM vs. Text-Promp for a variety of tasks on efficiency and performance. Note that we use full finetune for User-LLM. Relative improvement is calculated as (User-LLM - TP) / TP * 100\%.}
    \label{tab:next_item_single_modal}
\end{table*}

This subsection compares model accuracy and efficiency across various tasks and baselines. All reported \userllm experiments utilize pretrained encoders and finetune all three components (encoder, projection layers, and LLM) during cotraining, unless stated otherwise.

\subsubsection{Overall Performance} \userllm demonstrates significant efficiency gains (Table~\ref{tab:next_item_single_modal}), achieving a remarkable \textbf{12-21.9X} reduction in FLOPs compared to the Text Prompt baseline.  Moreover, \userllm consistently outperforms the Text Prompt baseline on most tasks across all three datasets, with performance improvements up to \textbf{16.33\%}. The exception is the Google review next item task, where \userllm exhibits slightly lower performance. 

While pretrained on next item prediction tasks, the user encoders generates embeddings encapsulating generalized user context, enabling \userllm to generalize well to diverse personalization tasks (e.g., favorite genre/categroy prediction, review generation) (Table~\ref{tab:next_item_single_modal}). These tasks require deep user understanding, demonstrating \userllm's effectiveness in understanding user preferences from interaction data.

\begin{figure}[t]
    \centering
    \includegraphics[width=1.0\columnwidth]{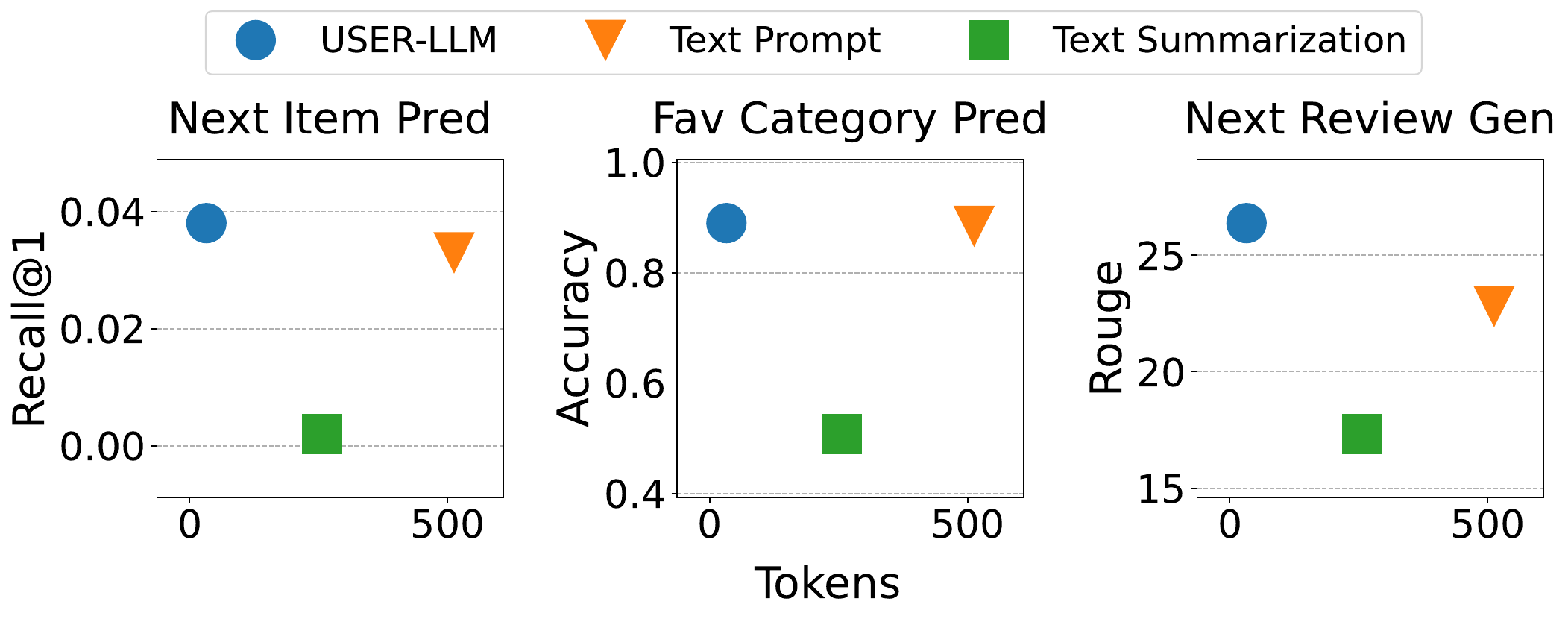}
    \caption{\userllm v.s. Text Prompt (TP) v.s. Text Summarization (TS) for 3 tasks on Amazon Review dataset. \userllm achieves better performance with much fewer tokens.}
    \label{fig:tp-userllm-pvt}
\end{figure}


Fig.~\ref{fig:tp-userllm-pvt} compares the performance of \userllm against Text Prompt (TP) and Text Summarization (TS) baselines on three tasks using the Amazon Review dataset: Next Item Prediction, Favorite Category Prediction, and Next Review Generation. The x-axis represents the number of tokens used, highlighting the efficiency of each method. Notably, \userllm consistently achieves superior performance while utilizing significantly fewer tokens. This efficiency is evident in all three tasks, indicating the effectiveness of directly embedding user history as a distinct modality.  \userllm requires fewer tokens to achieve equal or better performance, suggesting its ability to extract and leverage more relevant information from user data.

\subsubsection{Varying Context Length }  

To understand the impact of input length on performance and efficiency, we trained \userllm with varying predefined sequence lengths (50, 100, and 200 items) on MovieLens20M. Both frozen and tunable LLMs were evaluated and compared against a text-prompt baseline fine-tuned under the same computational constraints.

Table~\ref{tab:next-item-seqlen-flops} highlights the significant computational advantage of  \userllm. As the input sequence length increases, the required number of input tokens for the text-prompt approach scales proportionally (700, 1350, and 2500 tokens for lengths 50, 100, and 200 respectively).  This leads to a substantial increase in computational cost and memory demands.  In contrast, \userllm leverages a user encoder that compresses user activity sequences into compact embeddings. This allows the LLM to operate on a fixed-length input of only 32 tokens, regardless of the input sequence length. To illustrate the efficiency gain, we can approximate FLOPs using the heuristic  $FLOPs \approx 6ND$ \cite{kaplan2020scaling}, where $N$ represents the number of LLM parameters, and $D$ represents the total number of training tokens ($D = BSL$, with batch size $B$, training steps $S$, and input length $L$). Given that the encoder and \perceiver parameters ($N_e$ and $N_p$ respectively) are significantly smaller than the LLM parameters ($N_e \ll N$ and $N_p \ll N$), their computational costs are considered negligible. Using this approximation, \userllm achieves a remarkable FLOPs reduction compared to the text-prompt baseline, reaching up to \textbf{78.1X} when the input sequence length reaches 200.

\subsubsection{Frozen LLM }

Furthermore,  \userllm excels at integrating LLMs while preserving their inherent knowledge. By utilizing a training strategy (Enc) that keeps the LLM frozen, we prevent catastrophic forgetting and maintain the LLM's pretrained capabilities. As demonstrated in Table~\ref{tab:next-item-seqlen-flops}, \userllm with a frozen LLM surpasses the performance of text-prompt-based methods utilizing both full fine-tuning and LoRA tuning.  Impressively, it achieves comparable performance to \userllm  trained with an unfrozen LLM. These findings highlight \userllm's ability to effectively contextualize user preferences while safeguarding the LLM's pretrained knowledge, effectively mitigating the drawbacks associated with text-prompt-based methods.
\subsubsection{Parameter efficiency} \userllm requires fewer tunable parameters to achieve competitive performance. Table~\ref{tab:parameter_efficiency} shows the model accuracy across multiple training strategies. Notably, the \textit{Enc} approach (tuning encoder and projection layers and freezing the LLM) achieved comparable task accuracy to full finetuning while requiring significantly fewer tuned parameters. This is because it effectively leverages the pretrained weights of both the user encoder and LLM, minimizing the need for extensive parameter updates during training. Thus, our approach offers a parameter-efficient solution for personalizing LLMs, facilitating model tuning on resource-constrained devices.

\begin{table}[t]
    \centering
    \small
    \begin{tabular}{ccccccc}
        \toprule
        \textbf{Dataset} & \textbf{Task} & \textbf{Metric} & \textbf{Full}  & \textbf{LoRA} & \textbf{Enc} & \textbf{Proj} \\
        \midrule
        \multirow{2}{*}{\makecell{\scriptsize{MovieLens} \\ \scriptsize{20M}}} & \scriptsize{NextItem} & \scriptsize{Recall@1} & \scriptsize{\textbf{0.055}} & \scriptsize{0.051} & \scriptsize{0.050} & \scriptsize{0.023} \\ 
         & \scriptsize{FavGenre} & \scriptsize{Recall@1} & \scriptsize{\textbf{0.787}} & \scriptsize{0.787} & \scriptsize{0.786} & \scriptsize{0.743} \\ \hline
        \multirow{3}{*}{\makecell{\scriptsize{Google} \\ \scriptsize{Review}}} & \scriptsize{NextItem} & \scriptsize{Recall@1} & \scriptsize{0.015} & \scriptsize{\textbf{0.016}} & \scriptsize{0.015} & \scriptsize{0.015} \\ 
         & \scriptsize{FavCat} & \scriptsize{Recall@1} & \scriptsize{\textbf{0.862}} & \scriptsize{0.844} & \scriptsize{0.844} & \scriptsize{0.615} \\ 
         & \scriptsize{ReviewG} & \scriptsize{Rouge} & \scriptsize{11.72} & \scriptsize{11.64} & \scriptsize{\textbf{11.76}} & \scriptsize{9.61} \\ \hline
        \multirow{3}{*}{\makecell{\scriptsize{Amazon} \\ \scriptsize{Review}}} & \scriptsize{NextItem} & \scriptsize{Recall@1} & \scriptsize{\textbf{0.037}} & \scriptsize{0.028} & \scriptsize{0.028} & \scriptsize{0.014} \\ 
         & \scriptsize{FavCat} & \scriptsize{Recall@1} & \scriptsize{\textbf{0.890}} & \scriptsize{0.887} & \scriptsize{0.885} & \scriptsize{0.850} \\ 
         & \scriptsize{ReviewG} & \scriptsize{Rouge} & \scriptsize{\textbf{26.38}} & \scriptsize{24.56} & \scriptsize{24.30} & \scriptsize{19.04} \\ 
        \bottomrule
    \end{tabular}
    \caption{Comparison between different training strategies.}
    \label{tab:parameter_efficiency}
\end{table}

\begin{table}[t]
    \centering
    \small
    \begin{tabular}{ccccc}
        \toprule
         & & \textbf{Len50} & \textbf{Len100} & \textbf{Len200} \\
        \midrule
        \multirow{2}{*}{\# Tokens} & \userllm & 32 & 32 & 32 \\ 
        \cmidrule{2-5}
         & TP & 700 & 1350 & 2500 \\ 
        \midrule
        \multicolumn{2}{c}{FLOPs Reduction} & 21.9X & 42.2X & {\bf 78.1X}\\ \hline
        \multirow{2}{*}{\makecell{Performance\\improvement}} & Unfrozen LLM & 1.10X & 1.23X & {\bf 1.49X}\\ \cmidrule{2-5}
        & Frozen LLM & 2.78X & 2.97X & {\bf 3.27X}\\
        \bottomrule
    \end{tabular}
    \caption{LLM input token counts for \userllm v.s. TextPrompt(TP). FLOPs reduction and performance improvement achieved by \userllm. FLOPs reduction = FLOPs(TP) / FLOPs(\userllm). Performance improvement = Recall@1(\userllm) / Recall@1(TP).}
    \label{tab:next-item-seqlen-flops}
\end{table}



\begin{table}[t]
    \centering
    \small
    \begin{tabular}{ccccc}
        \toprule
         \multirow{2}{*}{\textbf{Task}} & \multicolumn{2}{c}{\makecell{\textbf{\userllm}}} & \multicolumn{2}{c}{\makecell{+\textbf{Perceiver}}} \\
        \cmidrule{2-5}
        & \textbf{Full} & \textbf{Enc} & \textbf{Full} & \textbf{Enc}\\
        \midrule
        NextItem & 0.055 & 0.050 & 0.054 & 0.054  \\ 
        FavGenre & 0.787 & 0.786 & 0.784 & 0.784  \\ 
        \hline
        \makecell[bc]{Tokens} & \multicolumn{2}{c}{50} & \multicolumn{2}{c}{16 (\textbf{-68\%})} \\ 
        \bottomrule
    \end{tabular}
    \caption{\userllm and its \perceiver variant effectively reduce the number of embedding tokens while maintaining competitive performance. Experiments are done on MovieLens 20M. Metric: Recall@1.}
    \label{tab:inference_efficiency}
\end{table}

\subsection{Ablation Study}
\label{sec:experiments-ablation}

\begin{figure}[t]
\centering
    \includegraphics[width=0.85\columnwidth]{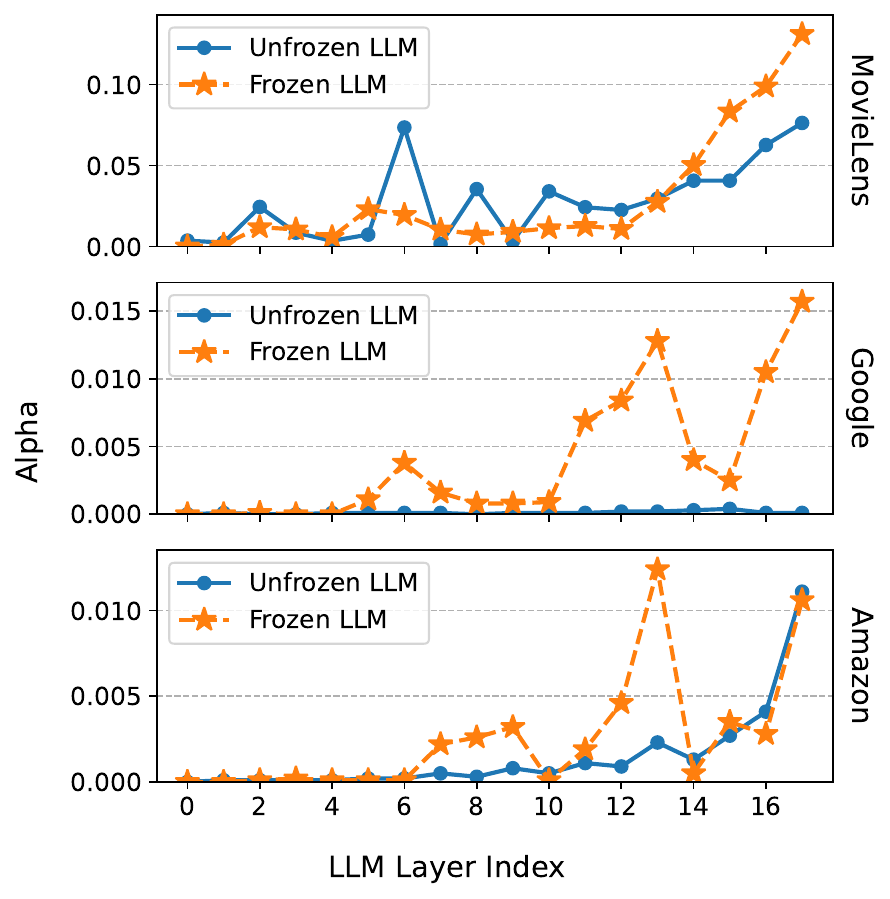}
    \caption{Converged gating parameters (Alpha) of different \crossattention layers indicate upper LLM transformer layers attend more signifcantly to encoder output than lower layers in next item prediction models on all three datasets.} 
    \label{fig:gated-alpha}
\end{figure}

\subsubsection{Benefit of pretraining}
We study the benefits of pretraining user encoders for downstream tasks. Results presented in columns 3 to 4 in Table~\ref{tab:pretrain-xatten-sp} demonstrate that models utilizing pretrained encoders consistently outperform those with randomly initialized encoders across all tasks, suggesting that pretraining enables encoders to capture user preference effectively, providing valuable context to LLMs during cotraining.

\subsubsection{Soft Prompt v.s. Cross Attention}
We compare the performance of \userllm with an alternative encoder-LLM integration strategy, \softprompt~\cite{lester-etal-2021-power}. We set the task prompt's length to $10$ and utilized a pretrained Autoregressive encoder in our experiment. As shown in the last column of Table~\ref{tab:pretrain-xatten-sp}, the cross-attention-based approach generally outperforms the soft-prompt approach. This advantage is particularly significant in the two review generation tasks, suggesting that the \crossattention mechanism enables LLMs to better extract and utilize the rich user information encoded within user embeddings. This highlights the value of cross-attention for tasks demanding a nuanced understanding of human intent.

\begin{table}[t]
    \centering
    \small
    \begin{tabular}{C{0.9cm}cccc}
        \toprule
        \multirow{3}{*}{\footnotesize{\textbf{Dataset}}} & \multirow{3}{*}{\footnotesize{\textbf{Task}}} & \multicolumn{2}{c}{\textbf{\crossattention}} & \multirow{2}{*}{\makecell{\textbf{\textit{soft}} \\ \textbf{\textit{prompt}}}}\\
        \cmidrule{3-4}
        & & \textbf{PT} & \textbf{RD}\\
        \midrule
        \multirow{2}{*}{\makecell[bc]{\scriptsize{MovieLens} \\ \scriptsize{20M}}} & \small{NextItem} & \textbf{0.055} & 0.043 & 0.047\\ 
         & \small{FavGenre} & \textbf{0.787} & 0.782 & \textbf{0.787}\\ \hline
        \multirow{3}{*}{\makecell{\scriptsize{Google} \\ \scriptsize{Review}}} & \small{NextItem} & \textbf{0.015} & \textbf{0.015} & 0.013 \\ 
         & \small{FavCat} & \textbf{0.862} & 0.853 & 0.822 \\ 
         & \small{ReviewG} & \textbf{11.72} & 11.43 & 9.06 \\ \hline
        \multirow{3}{*}{\makecell{\scriptsize{Amazon} \\ \scriptsize{Review}}} & \small{NextItem} & \textbf{0.037} & 0.026 & 0.031 \\ 
         & \small{FavCat} & 0.890 & 0.882 & \textbf{0.894} \\ 
         & \small{ReviewG} & \textbf{26.38} & 23.12 & 25.80 \\ 
        \bottomrule
    \end{tabular}
    \caption{Experiments on different model fusing architectures, encoders, and the benefit of pretraining. Report Recall@1 for next item prediction and favorite genre/category prediction, ROUGE-Lsum for review generation. Full finetuning is used throughout the experiments. \textbf{PT}: Pretrained encoder. \textbf{RD}: Randomly initialized encoder. \textbf{\softprompt}: Connect user embeddings with LLM via soft-prompt.}
    \label{tab:pretrain-xatten-sp}
\end{table}

\subsubsection{Gated Cross-attention}
We further assess \crossattention in \userllm by employing gated cross-attention layers that include a trainable gating parameter $\alpha$ governing the amount of cross-attention to be added to the language model stack. Fig. \ref{fig:gated-alpha} shows the converged values of gating parameters of 18 gated cross-attention layers for next item prediction tasks on MovieLens20M, Amazon Review and Google Local Review. The $\alpha$ values suggest that the upper LM layers attend more significantly to encoder output compared to the lower layers.  Since the encoder output is user embedding that typically captures high-level semantics such as user attributes and interests, our result suggests that the upper layers in \userllm are likely to process such semantic information, hence attending more to encoder output.  Interestingly, previous work by Geva and others \cite{geva-etal-2021-transformer} showed that lower layer keys in transformer correlate more with shallow surface features, whereas upper layer keys capture semantic topics. These analyses provide a plausible explanation why \crossattention outperforms \softprompt in \userllm.


\section{Conclusion and Future Work}

In this paper, we introduced \userllm, a framework for contextualizing LLMs through user embeddings. These embeddings, derived from self-supervised pretraining on diverse user interactions, capture hidden user preferences and their evolution. By integrating these embeddings with LLMs through \crossattention, \userllm empowers LLMs to adjust dynamically to user contexts.

Our comprehensive evaluation on MovieLens, Amazon Review, and Google Local Review datasets demonstrated that \userllm consistently outperforms text-prompt-based approaches on tasks requiring deep user understanding, particularly those involving long user interaction histories. \userllm's computational efficiency and ability to preserve LLM knowledge further make it particularly well-suited for practical applications where real-time personalization is crucial.

Future research directions include optimizing user embedding generation through advanced pretraining techniques and investigating the alignment between user embeddings and the language model for even deeper user context understanding. Additionally, training \userllm on a diverse range of tasks could further enhance its generalization abilities and adaptability across a broader spectrum of user scenarios. By addressing these aspects, \userllm has the potential to unlock the full power of LLMs for building truly personalized and adaptive AI agents.

\bibliography{references}

\newpage
\appendix

\section{More details about Datasets}

\begin{itemize}
\itemsep0.2em 
    \item \textbf{MovieLens20M\footnote{https://grouplens.org/datasets/movielens/}} A widely used benchmark dataset \cite{Harper2015MovieLens} for evaluating personalization and recommendation algorithms. Interaction modalities are movie name, genre, and rating. 
    \item \textbf{Google Local Review Dataset\footnote{https://datarepo.eng.ucsd.edu/mcauley\_group/gdrive/googlelocal/}} \cite{Li2022ACL,Yan2023SIGIR} This dataset contains review information on Google Maps up to Sep 2021. We used New York's data in our experiments. Interaction modalities are place name, place category, rating, and review.
    \item \textbf{Amazon Review Dataset\footnote{https://nijianmo.github.io/amazon/index.html} }\cite{amazon_data_2016} This is a series of product review datasets crawled from Amazon.com. We used the "Movie\_and\_TV" category for evaluation. Interaction modalities are product title, product category, rating, and review summary (a short opinion summary written by the user).
\end{itemize}

\section{Training Hyperparameters}
In Table~\ref{tab:training_hparams}, we show training hyperparameters of \userllm, including learning rate, weight decay, batch size, and number of training steps. In addition, \userllm used the cosine learning rate decay across all experiments and used the first $10\%$ of training steps for linear warmup.
\begin{table*}[ht]
    \centering
    \small
    \begin{tabularx}{\textwidth}{Y|Y|Y|Y|Y|Y}
        \toprule
        \textbf{Dataset} & \textbf{Task} & \textbf{Learning rate} & \textbf{Weight decay} & \textbf{Batch size} & \textbf{Training steps}\\
        \midrule
        \multirow{3}{*}{MovieLens 20M} & \small{EncoderPretrain} & 1e-1 & 1e-1 & 32,768 & 10,000 \\\cmidrule{2-6}
        & \small{NextItem} & 1e-3 & 1e-3 & 8,192 & 10,000\\ 
         & \small{FavGenre} & 1e-3 & 1e-3 & 1,024 & 5,000\\ \midrule
        \multirow{4}{*}{Google Review} & \small{EncoderPretrain} & 1e-4 & 1e-1 & 16,384 & 10,000 \\\cmidrule{2-6}
         & \small{NextItem} & 1e-3 & 1e-1 & 2,048 & 5,000\\ 
         & \small{FavCat} & 1e-3 & 1e-1 & 2,048 & 5,000 \\ 
         & \small{ReviewG} & 1e-3 & 1e-1 & 1,024 & 5,000 \\ \midrule
        \multirow{4}{*}{Amazon Review} & \small{EncoderPretrain} & 1e-4 & 1e-1 & 12,800 & 7,000 \\\cmidrule{2-6}
         & \small{NextItem} & 1e-3 & 1e-1 & 8,192 & 10,000 \\ 
         & \small{FavCat} & 1e-3 & 1e-1 & 1,024 & 5,000 \\ 
         & \small{ReviewG} & 1e-3 & 1e-1 & 2,048 & 5,000 \\ 
        
        \bottomrule
    \end{tabularx}
    \caption{Training hyperparameters.}
    \label{tab:training_hparams}
\end{table*}

\section{Encoder Architectures}
\begin{figure*}[th]
\centering
\includegraphics[width=0.9\textwidth]{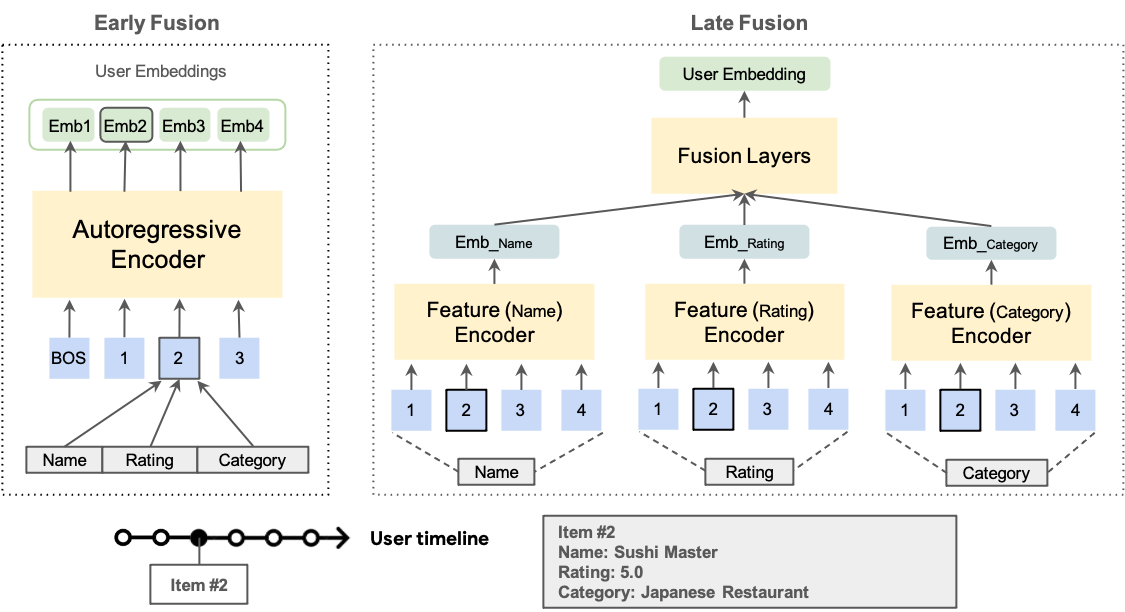}
    \caption{Encoder architectures supported in \userllm: \earlyfusion and \latefusion user encoders.}
    \label{fig:encoder-architectures}
\end{figure*}

\userllm uses a Transformer-based user encoder to generate user embeddings from ID-based feature sequences. 
To effectively fuse information from user interaction data, we explore and compare two fusion architectures: \earlyfusion and \latefusion.
Following the example shown in Fig.~\ref{fig:encoder-architectures}, each item in the user interaction timeline has three features: name, rating, and category. Each feature has its own vocabulary. Items from each feature are mapped to an integer ID and have their own embedding representations. In an \earlyfusion architecture, the $i_{th}$ item's three features (name, rating, and category) are combined into a fused embedding $f_i$. The sequence of fused embeddings is then fed into the user encoder to generate user embeddings. On the other hand, a \latefusion architecture employs three separate feature encoders to process each feature independently. The resulting embeddings are combined to a single user embedding at a later stage using fusion layers.

\textbf{Autoregressive Transformer} As described in Section~\ref{sec:embedding-gen}, \userllm employs an Autoregressive Transformer as the \earlyfusion encoder. Illustrated in Fig.~\ref{fig:user-llm}, this model receives a sequence of user activities, where each activity is represented by multiple features (name, rating, and category). Individual features are first embedded, then concatenated and processed by a Transformer decoder. The Autoregressive Transformer Encoder generates a user embedding for each input item in the user's timeline.

\textbf{Dual Encoder} The late-fusion encoder we investigated is a dual encoder architecture with separate 'user' and 'label' towers. The 'user tower' utilizes Transformers and processes a series of input tokens, while the 'label tower' processes the subsequent token (label) within the sequence. Each encoder generates an embedding representation of its respective input, and the similarity between these embeddings is calculated for loss computation via a softmax layer. When integrating with LLMs, we utilize only the 'user tower' for user embedding generation. With the default setting, a Dual Encoder generates a single user embedding from an input sequence. The number of generated embeddings can be adjusted through the configuration of the fusion layers.

Autoregressive Transformers and Dual Encoders offer different advantages for user embedding generation. Autoregressive Transformers excel at capturing long-range dependencies and contextual relationships within sequential data. However, they produce a fixed number of output embeddings (equal to the input sequence length), potentially introducing inefficiencies in the later LLM integration stage, especially when the sequence length is long. Furthermore, training these models can be computationally intensive. 

Compared to Autoregressive Transformers, Dual Encoders offer several advantages. The separate feature encoders within user towers allow independent processing of features, handling features with varying sequence lengths or misaligned timestamps and potentially enhancing information extraction. The configurable fusion layers provide flexibility in the number of generated embeddings. Additionally, Dual Encoders support diverse pretraining tasks beyond next item prediction, as user and label towers can process different features (e.g., name for the user tower and category for the label tower). However, Dual Encoders may be less effective at modeling the complex sequential patterns often present in user timelines. The optimal encoder choice depends on dataset characteristics, the desired level of contextual understanding, and a balance between computational cost and potential performance gain. 

Table~\ref{tab:user-llm-train-strategy-ar-de} compares the performance of using Autoregressive Transformer (AR) and Dual Encoder (DE) on three datasets and eight tasks, without computation constraint. AR consistently outperforms DE, across various training strategies and using both pretrained or randomly initialized encoders. Table~\ref{tab:ar-de-short-text} further analyzes the two encoders' performance when incorporating short-term history into the LLM prompt.

\section{Integration Approaches}
As mentioned in \ref{subsec:performance-efficiency-comp}, we explored the integration of user embeddings and LLMs in \userllm through two approaches: \crossattention and \softprompt. In the cross-attention method, the output embeddings from the pretrained user encoder are cross-attended with the LLM's intermediate text representations, similar to how Flamingo~\cite{alayrac2022flamingo} works. On the other hand, the \softprompt approach leverages user embeddings from the pretrained encoder as a soft prompt for the LLM. This soft prompt is prepended to the regular text prompt embedding, providing additional context about the user's interests and preferences. 

Table~\ref{tab:user-llm-train-strategy-xatten-sp} presents a detailed comparison between \crossattention and \softprompt approaches with various training strategies across all datasets and tasks.

\begin{table}[t]
    \centering
    \small
    \begin{tabular}{clp{0.7cm}cp{0.7cm}c}
        \toprule
        \multirow{2}{*}{\textbf{Dataset}} &  \multirow{2}{*}{\textbf{Recall}} & \multicolumn{2}{c}{\textbf{TextPrompt}} & \multicolumn{2}{c}{\textbf{\userllm}} \\
        \cmidrule{3-6}
        & & \textbf{TP10} & \textbf{TP50} & \textbf{AR50} & \textbf{AR50TP10} \\
        \midrule
        \multirow{3}{*}{\makecell[c]{\scriptsize{MovieLens} \\ \scriptsize{20M}}} & @1 & 0.049 & 0.049 & 0.055 & \textbf{0.059}  \\ 
         & @5 & 0.142 & 0.140 & 0.154 & \textbf{0.173}  \\ 
         & @10 & 0.211 & 0.206 & 0.243 & \textbf{0.252} \\ \hline
        \multirow{3}{*}{\makecell{\scriptsize{Google} \\ \scriptsize{Review}}} & @1 & \textbf{0.021} & \textbf{0.021} & 0.015 & \textbf{0.021} \\ 
         & @5 & 0.060 & \textbf{0.061} & 0.052 & \textbf{0.061} \\ 
         & @10 & 0.082 & \textbf{0.086} & 0.071 & 0.082 \\ \hline
        \multirow{3}{*}{\makecell{\scriptsize{Amazon} \\ \scriptsize{Review}}} & @1 & 0.038 & 0.034 & 0.037 & \textbf{0.041} \\ 
         & @5 & 0.047 & 0.042 & 0.047 & \textbf{0.051} \\ 
         & @10 & 0.050 & 0.047 & 0.051 & \textbf{0.055} \\ 
        \bottomrule
    \end{tabular}
    \caption{Experiments on short-term text prompt plus long-term embedding. \textbf{TP10}: TextPrompt baseline with $10$ recent user events. \textbf{TP50}: TextPrompt baseline with $50$ recent user events. \textbf{AR50}: With $50$ user events as encoder inputs. \textbf{AR50TP10}: With $50$ user events as encoder inputs and $10$ most recent events in text format as LLM inputs.}
    \label{tab:encoder-short-text}
\end{table}

\begin{table*}[h]
    \centering
    \small
    \begin{tabular}{p{0.2cm}p{0.8cm}p{0.8cm}C{0.58cm}C{0.58cm}C{0.58cm}C{0.58cm}C{0.58cm}C{0.58cm}C{0.58cm}C{0.58cm}C{0.58cm}C{0.58cm}C{0.58cm}C{0.58cm}C{0.58cm}C{0.58cm}}
        \toprule
        & \multirow{4}{*}{\small{Dataset}} & \multirow{4}{*}{Metric} & \multicolumn{8}{c}{AREnc-Xatten(\userllm)} & \multicolumn{6}{c}{DualEnc-Xatten} \\
        \cmidrule(lr){4-11} \cmidrule(lr){12-17}
        & & & \multicolumn{2}{c}{Full} & \multicolumn{2}{c}{LoRA} & \multicolumn{2}{c}{Enc} &  \multicolumn{2}{c}{Proj} & \multicolumn{2}{c}{Full} & \multicolumn{2}{c}{Enc} &  \multicolumn{2}{c}{Proj} \\
        \cmidrule(lr){4-11} \cmidrule(lr){12-17}
        & & & PT & RD & PT & RD & PT & RD & PT & RD & PT & RD & PT & RD & PT & RD \\
        \midrule
        \multirow{9}{*}{\makecell[bl]{\rotatebox{90}{\scriptsize{Next Item}}}} & \multirow{3}{*}{\makecell[bl]{\scriptsize{MovieLens} \\ \scriptsize{20M}}} & R@1 & \textbf{0.055} & 0.041 & \underline{0.051} & 0.043 & 0.050 & 0.041 & 0.023 & 0.002 & 0.044 & 0.025 & 0.034 & 0.010 & 0.032 & 0.001 \\ 
        & & R@5 & 0.154& 0.133 & \textbf{0.164}  & 0.136 & \textbf{0.164} & 0.133 & 0.134 & 0.008 & 0.142 & 0.102 & 0.116 & 0.043 & 0.107 & 0.008 \\ 
        & & R@10 & \underline{0.243} & 0.203 & \textbf{0.244} & 0.205 & \underline{0.243} & 0.204 & 0.203 & 0.013 & 0.218 & 0.169 & 0.182 & 0.076 & 0.166 & 0.012 \\ \cmidrule{2-17}
        & \multirow{3}{*}{\makecell[bc]{\scriptsize{Google} \\ \scriptsize{Review}}} & R@1 & \underline{0.015} & \underline{0.015} & \textbf{0.016} & \underline{0.015} & \underline{0.015} & \underline{0.015} & \underline{0.015} & \underline{0.015} & 0.014 & \underline{0.015} & 0.014 & \underline{0.015} & \underline{0.015} & \underline{0.015} \\ 
        & & R@5 & \textbf{0.052} & \textbf{0.052} & 0.051 & 0.050 & \underline{0.051} & \underline{0.051} & 0.050 & 0.042 & 0.046 & 0.044 & 0.047 & 0.049 & 0.049 & 0.044 \\ 
        & & R@10 & \textbf{0.071} & \textbf{0.071} & 0.070 & 0.070 & 0.070 & 0.070 & 0.070 & 0.057 & 0.066 & 0.063 & 0.061 & 0.067 & 0.064 & 0.057 \\ \cmidrule{2-17}
        & \multirow{3}{*}{\makecell[bc]{\scriptsize{Amazon} \\ \scriptsize{Review}}} & R@1 & \textbf{0.037} & 0.023 & \underline{0.028} & 0.018 & \underline{0.028} & 0.022 & 0.014 & 0.000 & 0.020 & 0.005 & 0.016 & 0.002 & 0.008 & 0.000 \\ 
        & & R@5 & \textbf{0.047} & 0.031 & \underline{0.035} & 0.026 & 0.034 & 0.029 & 0.019 & 0.000 & 0.024 & 0.010 & 0.019 & 0.005 & 0.010 & 0.002 \\ 
        & & R@10 & \textbf{0.051} & 0.037 & \underline{0.038} & 0.029 & 0.036 & 0.030 & 0.021 & 0.0003 & 0.026 & 0.014 & 0.019 & 0.009 & 0.011 & 0.003 \\ 
        \midrule
        \multirow{9}{*}{\makecell[bl]{\rotatebox{90}{\scriptsize{Fav Genre/Category}}}} & \multirow{3}{*}{\makecell[bl]{\scriptsize{MovieLens} \\ \scriptsize{20M}}} & R@1 & \textbf{0.787} & 0.778 & \textbf{0.787} & 0.779 & 0.786 & 0.778 & 0.743 & 0.484 & 0.775 & 0.704 & 0.776 & 0.759 & 0.649 & 0.484 \\ 
        & & R@5 & \textbf{0.996} & 0.994 & \textbf{0.996} & 0.994 & \textbf{0.996} & 0.994 & 0.993 & 0.931 & 0.994 & 0.959 & 0.993 & 0.980 & 0.976 & 0.931 \\ 
        & & R@10 & 0.999 & 0.999 & 0.999 & 0.999 & 0.999 & 0.999 & 0.999 & 0.991 & 0.999 & 0.987 & 0.999 & 0.997 & 0.998 & 0.991 \\ \cmidrule{2-17}
        & \multirow{3}{*}{\makecell[bc]{\scriptsize{Google} \\ \scriptsize{Review}}} & R@1 & \textbf{0.862} & \underline{0.853} & 0.844 & 0.785 & 0.844 & 0.775 & 0.615 & 0.142 & 0.797 & 0.642 & 0.760 & 0.508 & 0.271 & 0.142 \\ 
        & & R@5 & \textbf{0.951} & \underline{0.944} & 0.938 & 0.915 & 0.937 & 0.911 & 0.855 & 0.505 & 0.920 & 0.842 & 0.885 & 0.711 & 0.607 & 0.505 \\ 
        & & R@10 & \textbf{0.964} & \underline{0.956} & 0.948 & 0.941 & 0.949 & 0.936 & 0.909 & 0.660 & 0.944 & 0.871 & 0.914 & 0.798 & 0.763 & 0.661 \\ \cmidrule{2-17}
        & \multirow{3}{*}{\makecell[bc]{\scriptsize{Amazon} \\ \scriptsize{Review}}} & R@1 & \textbf{0.890} & 0.882 & \underline{0.887} & 0.884 & 0.885 & 0.886 & 0.850 & 0.223 & 0.854 & 0.745 & 0.843 & 0.714 & 0.312 & 0.223 \\ 
        & & R@5 & \underline{0.934} & \textbf{0.948} & 0.917 & 0.931 & 0.917 & 0.930 & 0.918 & 0.663 & 0.912 & 0.870 & 0.898 & 0.866 & 0.694 & 0.662 \\ 
        & & R@10 & \underline{0.944} & \textbf{0.958} & 0.919 & 0.934 & 0.919 & 0.932 & 0.923 & 0.793 & 0.934 & 0.925 & 0.930 & 0.903 & 0.846 & 0.793 \\ 
        \midrule
        \multirow{2}{*}{\makecell[bl]{\rotatebox{90}{\scriptsize{Review Gen}}}} & \scriptsize{Google Review} & Rouge & \underline{11.72} & 11.43 & 11.64 & 11.55 & \textbf{11.76} & 11.56 & 9.61 & 8.05 & 11.71 & 11.04 & \textbf{11.76} & 10.82 & 9.18 & 6.62 \\ 
        \cmidrule{2-17}
        & \scriptsize{Amazon Review} & Rouge & \textbf{26.38} & 23.12 & 24.56 & 25.00 & 24.30 & \underline{25.18} & 19.04 & 17.30 & 23.83 & 18.1 & 21.11 & 20.78 & 18.92 & 18.53 \\ 
        \bottomrule
    \end{tabular}
    \caption{\userllm experiments results with different training strategies and different user encoder architectures. \textbf{PT}: Pretrained user encoder. \textbf{RD}: Randomly initialized user encoder.}
    \label{tab:user-llm-train-strategy-ar-de}
\end{table*}

\begin{table*}[h]
    \centering
    \small
    \begin{tabular}{clp{0.7cm}cp{0.7cm}c}
        \toprule
        \multirow{2}{*}{\small{\textbf{Dataset}}} &  \multirow{2}{*}{\small{\textbf{Recall}}} & \multicolumn{2}{c}{\small{\textbf{AREnc-Xatten}}} & \multicolumn{2}{c}{\small{\textbf{DualEnc-Xatten}}} \\
        \cmidrule{3-6}
        & & \small{\textbf{AR50}} & \small{\textbf{AR50TP10}} & \small{\textbf{DE50}} & \small{\textbf{DE50TP10}} \\
        \midrule
        \multirow{3}{*}{\makecell[c]{\scriptsize{MovieLens} \\ \scriptsize{20M}}} & @1 & \underline{0.055} & \textbf{0.059} & 0.044 & 0.051  \\ 
         & @5 & \underline{0.164} & \textbf{0.173} & 0.135 & 0.151  \\ 
         & @10 & \underline{0.243} & \textbf{0.252} & 0.206 & 0.221 \\ \hline
        \multirow{3}{*}{\makecell{\scriptsize{Google} \\ \scriptsize{Review}}} & @1 & 0.015 & \textbf{0.021} & 0.014 & \underline{0.020} \\ 
         & @5 & 0.052 & \textbf{0.061} & 0.046 & \underline{0.055} \\ 
         & @10 & 0.071 & \textbf{0.082} & 0.066 & \underline{0.078} \\ \hline
        \multirow{3}{*}{\makecell{\scriptsize{Amazon} \\ \scriptsize{Review}}} & @1 & \underline{0.037} & \textbf{0.041} & 0.020 & 0.030 \\ 
         & @5 & \underline{0.047} & \textbf{0.051} & 0.024 & 0.038 \\ 
         & @10 & \underline{0.051} & \textbf{0.055} & 0.026 & 0.040 \\ 
        \bottomrule
    \end{tabular}
    \caption{Experiments on short-term text prompt long-term embedding with two encoder architectures: Autoregressive Encoder and Dual Encoder. \textbf{AR50}: With $50$ user events as Autoregressive Encoder inputs. \textbf{AR50TP10}: With $50$ user events as Autoregressive Encoder inputs and $10$ most recent events in text format as LLM inputs. \textbf{DE50}: With $50$ user events as Dual Encoder inputs. \textbf{DE50TP10}: With $50$ user events as Dual Encoder inputs and $10$ most recent events in text format as LLM inputs.}
    \label{tab:ar-de-short-text}
\end{table*}

\begin{table*}[h]
    \centering
    \small
    \begin{tabular}{lllcccccccc}
        \toprule
        & \multirow{3}{*}{\small{Dataset}} & \multirow{3}{*}{\small{Metric}} & \multicolumn{4}{c}{\crossattention(\userllm)} & \multicolumn{4}{c}{\softprompt} \\
        \cmidrule(lr){4-7} \cmidrule(lr){8-11}
        & & & Full & LoRA & Enc & Proj & Full & LoRA & Enc & Proj\\
        \midrule
        \multirow{3}{*}{\makecell[bc]{NextItem}} & \scriptsize{MovieLens20M} & R@1 & \textbf{0.055} & \underline{0.051} & 0.050 & 0.023 & 0.051 & 0.036 & 0.0047 & 0.0025 \\ 
        & \scriptsize{Google Review} & R@1 & \underline{0.015} & \textbf{0.016} & \underline{0.015} & \underline{0.015} & 0.013 & 0.008 & 0.012 & 0.008 \\ 
        & \scriptsize{Amazon Review} & R@1 & \textbf{0.037} & 0.028 & 0.028 & 0.014 & \underline{0.031} & 0.010 & 0.003 & 0.003 \\ 
        \midrule
        \multirow{3}{*}{\makecell[bc]{FavGenre\\(FavCat)}} & \scriptsize{MovieLens20M} & R@1 & 0.787 & 0.787 & 0.786 & 0.743 & \textbf{0.787} & \textbf{0.787} & 0.781 & 0.692 \\ 
        & \scriptsize{Google Review} & R@1 & \textbf{0.862} & \underline{0.844} & \underline{0.844} & 0.615 & 0.822 & 0.729 & 0.694 & 0.259 \\ 
        & \scriptsize{Amazon Review} & R@1 & \underline{0.890} & 0.887 & 0.885 & 0.850 & \textbf{0.894} & 0.877 & 0.831 & 0.356 \\ 
        \midrule
        \multirow{2}{*}{\makecell[bc]{ReviewG}} & \scriptsize{Google Review} & ROUGE & \underline{11.72} & 11.64 & \textbf{11.76} & 9.61 & 9.06 & 9.04 & 8.75 & 7.76 \\
        & \scriptsize{Amazon Review} & ROUGE & \textbf{26.38} & 24.56 & 24.30 & 19.04 & \underline{25.80} & 22.86 & 20.36 & 15.12 \\ 
        \bottomrule
    \end{tabular}
    \caption{\userllm experiments results for \crossattention \& \softprompt with different training strategies (with pretrained Autoregressive encoders). Report Recall@1 for next item prediction and favorite genre/category prediction, ROUGE-Lsum for review generation.}
    \label{tab:user-llm-train-strategy-xatten-sp}
\end{table*}

\section{More Results}

\begin{table*}[h]
    \centering
    \small
    \begin{tabular}{l|>{\centering}p{0.52cm}c|>{\centering}p{0.52cm}c|>{\centering}p{0.52cm}c||>{\centering}p{0.52cm}c|>{\centering}p{0.52cm}c|>{\centering}p{0.52cm}c}
        \toprule
        \multirow{3}{*}{\textbf{Recall}} & \multicolumn{6}{c||}{\textbf{Unfrozen LLM}} & \multicolumn{6}{c}{\textbf{Frozen LLM}} \\
        \cline{2-13}
        & \multicolumn{2}{c|}{\textbf{Len50}} & \multicolumn{2}{c|}{\textbf{Len100}} & \multicolumn{2}{c||}{\textbf{Len200}} & \multicolumn{2}{c|}{\textbf{Len50}} & \multicolumn{2}{c|}{\textbf{Len100}} & \multicolumn{2}{c}{\textbf{Len200}} \\
        \cline{2-13}
        & \scriptsize{\textbf{TP}} & \scriptsize{\textbf{\userllm}} & \scriptsize{\textbf{TP}} & \scriptsize{\textbf{\userllm}} & \scriptsize{\textbf{TP}} & \scriptsize{\textbf{\userllm}} & \scriptsize{\textbf{TP}} & \scriptsize{\textbf{\userllm}} & \scriptsize{\textbf{TP}} & \scriptsize{\textbf{\userllm}} & \scriptsize{\textbf{TP}} & \scriptsize{\textbf{\userllm}} \\
        \midrule
        @1 & 0.049 & \textbf{0.055} & 0.043 & \textbf{0.055} & 0.034 & \textbf{0.055}  & 0.017 & \textbf{0.051} & 0.016 & \textbf{0.051} & 0.015 & \textbf{0.053} \\ 
        @5 & 0.140 & \textbf{0.154} & 0.123 & \textbf{0.153} & 0.102 & \textbf{0.155} & 0.051 & \textbf{0.147} & 0.046 & \textbf{0.146} & 0.042 & \textbf{0.146} \\ 
        @10 & 0.206 & \textbf{0.227} & 0.183 & \textbf{0.226}  & 0.154 & \textbf{0.229} & 0.077 & \textbf{0.214} & 0.072 & \textbf{0.214} & 0.066 & \textbf{0.216} \\ \hline
        \bottomrule
    \end{tabular}
    \caption{Varying sequence length experiments on MovieLens20M with frozen LLM or unfrozen LLM. \textbf{Unfrozen LLM}: full finetune for TextPrompt(TP) model, training strategy \textit{\textbf{Full}} for \userllm. \textbf{Frozen LLM}: LoRA parameters tuning for TextPrompt(TP) model, training strategy \textit{\textbf{Enc}} for \userllm.}.
    \label{tab:long-sequence-all-recalls}
\end{table*}

\subsection{Autoregressive Encoder v.s. Dual Encoder}
Table~\ref{tab:ar-de-short-text} compares Autoregressive Encoder and Dual Encoder on long-term and short-term user context tasks. Refer to Columns 4 and 12 in Table~\ref{tab:user-llm-train-strategy-ar-de} for the performance of these two encoders when utilizing a pretrained encoder and using \textbf{Full} training strategy.

\subsection{Training Strategies and Encoder Pretraining}
Table~\ref{tab:user-llm-train-strategy-ar-de} shows the performance of Autoregressive encoder based \userllm under different training strategies. It also compares results between using a pretrained user encoder and a randomly initialized user encoder in the encoder-LLM integration stage.

\subsection{Cross-attention v.s. Soft-prompt}
Table~\ref{tab:user-llm-train-strategy-xatten-sp} compares the performance of two encoder-LLM integration strategies: \crossattention and \softprompt.

\subsection{Next Item Prediction}
Table~\ref{tab:long-sequence-all-recalls} show results comparing text-prompt-based approach and \userllm in next item prediction task on MovieLens20M dataset with varying input activity sequence length.

\subsection{Long-term and short-term user context}
\userllm can integrate both long-term and short-term user context. We can feed long-term user history into the user encoder to generate compact user embeddings while prompting LLM with short-term user interactions. As shown in Table~\ref{tab:encoder-short-text}, \userllm generally achieves better results when combining long-term and short-term user context for next item prediction tasks. Using 50 user activities for embedding generation and 10 most recent activities for prompting LLM, \userllm (\textit{AR50TP10}) outperforms TextPrompt with only short-term context (\textit{TP10}) by providing long-term user information and outperforms TextPrompt with long-term context (\textit{TP50}) by extracting meaningful user representations efficiently. Compared to \textit{AR50}, in which \userllm only uses long-term user data, short-term context provides the model (\textit{AR50TP10}) with additional natural language information about item names. This leads to improved next item name prediction while preserving computational efficiency.

\section{Review Generation Examples}
We presented some examples from Amazon/Google review generation tasks below. From these examples, we can observe personalized responses generated using \userllm exhibit greater alignment with user preferences and backgrounds compared to unpersonalized responses. While discrepancies remain between personalized responses using \userllm and the ground truth, \userllm demonstrates the ability to produce reviews that reflect specific user characteristics and inclinations more effectively.

\begin{itemize}
    \item \textbf{Input prompt:} ``Write a review for The Firm - Aerobic Body Shaping."
        \SubItem{\textbf{Ground truth response:} ``Save Your Money."}
        \SubItem{\textbf{Personalized response:} ``Terrible Waste of Time."}
        \SubItem{\textbf{Unpersonalized response:} ``Aerobic Workout."}
    \item \textbf{Input prompt:} ``Write a review for A.I. - Artificial Intelligence."
        \SubItem{\textbf{Ground truth response:} ``A pure movie masterpiece of a film by Steven Speilberg.  You will either love or hate this movie, but it will make you think."}
        \SubItem{\textbf{Personalized response:} ``If you are into science fiction and you are looking for a movie that is not only good but also educational and fun, 'A.I."}
        \SubItem{\textbf{Unpersonalized response:} ``A.I. - Artificial Intelligence."}
    \item \textbf{Input prompt:} ``Write a review for The Family that Preys."
        \SubItem{\textbf{Ground truth response:} ``Four Stars."}
        \SubItem{\textbf{Personalized response:} ``Four Stars."}
        \SubItem{\textbf{Unpersonalized response:} ``A great thriller!"}
    \item \textbf{Input prompt:} ``Write a review for Green Street Hooligans 2."
        \SubItem{\textbf{Ground truth response:} ``Hmmmmmm..."}
        \SubItem{\textbf{Personalized response:} ``Not for everyone."}
        \SubItem{\textbf{Unpersonalized response:} ``A good sequel."}
    \item \textbf{Input prompt:} ``Help me write a review for Blue with a rating of 5.0."
        \SubItem{\textbf{Ground truth response:} ``(Translated by Google) Very delicious and the service staff are very enthusiastic."}
        \SubItem{\textbf{Personalized response:} ``(Translated by Google) Good food."}
        \SubItem{\textbf{Unpersonalized response:} ``Great food and service."}
    \item \textbf{Input prompt:} ``Help me write a review for The Home Depot with a rating of 2.0."
        \SubItem{\textbf{Ground truth response:} ``(Translated by Google) The service is terrible (Original) El servicio esta pésimo."}
        \SubItem{\textbf{Personalized response:} ``(Translated by Google) Very bad service (Original) Muy mal servicio."}
        \SubItem{\textbf{Unpersonalized response:} ``No help."}

\end{itemize}

\end{document}